%% file: main.tex
\documentclass[11pt,a4paper]{article}
\pdfoutput=1

\usepackage[hyperref]{acl2018}
\usepackage{times}
\usepackage{url}
\usepackage{latexsym}
\usepackage[pdftex]{graphicx}
\usepackage[T1]{fontenc}
\usepackage[utf8]{inputenc}
\usepackage[export]{adjustbox}
\usepackage[capitalise]{cleveref}
\usepackage{booktabs}
\usepackage{csquotes}
\usepackage{pgf-pie}

\usepackage{pgfplots}
\pgfplotsset{compat=1.14}
\pgfplotsset{scaled ticks=false}

\usepackage{floatrow}
\setkeys{Gin}{width=0.85\linewidth}

\graphicspath{{img/}}

\title{A Crowd-Annotated Spanish Corpus for Humor Analysis}

\author{Santiago Castro, Luis Chiruzzo, Aiala Rosá, Diego Garat, Guillermo Moncecchi \\
Grupo de Procesamiento de Lenguaje Natural \\
Facultad de Ingeniería \\
Universidad de la República --- Uruguay \\
{\tt \{sacastro,luischir,aialar,dgarat,gmonce\}@fing.edu.uy} \\}

\aclfinalcopy{}

\begin{document}

\maketitle

\begin{abstract}
Computational Humor involves several tasks, such as humor recognition, humor generation, and humor scoring, for which it is useful to have human-curated data. In this work we present a corpus of 27,000 tweets written in Spanish and crowd-annotated by their humor value and funniness score, with about four annotations per tweet, tagged by 1,300 people over the Internet. It is equally divided between tweets coming from humorous and non-humorous accounts. The inter-annotator agreement Krippendorff's alpha value is 0.5710. The dataset is available for general use and can serve as a basis for humor detection and as a first step to tackle subjectivity.
\end{abstract}

\input{intro}
\input{extraction}
\input{annotation}
\input{corpus}
\input{analysis}
\input{conclusion}

\section*{Acknowledgments}

We thank everyone who annotated tweets via the web page. We would not have been able to reach the large number of annotations we have got without their help.

\bibliography{biblio}
\bibliographystyle{acl_natbib}

\end{document}

%% file: intro.tex
\section{Introduction}

Computational Humor studies humor from a computational perspective, involving several tasks such as humor recognition, which aims to tell if a piece of text is humorous or not; humor generation, with the objective of generating new texts with funny content; and humor scoring, whose goal is to predict how funny a piece of text is.

In order to carry out this kind of tasks through supervised machine learning methods, human-curated data is necessary. \citet{castro2016joke} built a humor classifier for Spanish and provided a dataset for humor recognition. However, there are some issues: few annotations per instance, low annotator agreement, and limited variety of sources for the humorous and mostly for the non-humorous tweets (the latter were only about news, inspirational thoughts and curious facts). Up to our knowledge, there is no other dataset to work on humor comprehension in Spanish. Some other authors, such as \citet{mihalcea2005bootstrapping, Mihalcea:2005:MCL:1220575.1220642, conf/wilf/SjoberghA07} have tackled humor recognition in English texts, building their own corpora by downloading \emph{one-liners} (one-sentence jokes) from the Internet, since working with longer texts would involve additional work, such as determining humor scope.

The microblogging platform Twitter has been found particularly useful for building humor corpora due to its public availability and the fact that its short messages are suitable for jokes or humorous comments. \citet{castro2016joke} built their corpus based on Twitter, selecting nine humorous accounts and nine non-humorous accounts about news, thoughts and curious facts. \citet{reyes2013multidimensional} built a corpus for detecting irony in tweets by searching for several hashtags (i.e., \#irony, \#humor, \#education and \#politics), which is also used in \citet{barbieri2014automatic} to train a classifier that detects humor. More recently, \citet{potash2017semeval} built a corpus based on tweets that aims to distinguish the degree of funniness in a given tweet. They used the tweet set issued in response to a TV game show, labeling which tweets were considered humorous by the show.

In this work we present a crowd-annotated Spanish corpus of tweets tagged with a humor/no humor value and also by a funniness score from one to five. The corpus contains tweets extracted from varied sources and has several annotations per tweet, reaching a high humor inter-annotator agreement.

The contribution of this work is twofold: the dataset is not only useful for building a humor classifier but it also serves to approach subjectivity in humor and funniness. Even though there are not enough  annotations per tweet as required to study subjectivity in a genuine way with techniques such as the ones by \citet{geng2016label}, the dataset aids as a playground to study the funniness and disagreement among several people.

This document is organized as follows. \cref{sec:extraction} explains where and how we obtained the data, and \cref{sec:annotation} describes how it was annotated. In \cref{sec:corpus} we present the corpus, and we address the analysis in \cref{sec:analysis}. Finally, in \cref{sec:conclusion} we present draw the conclusions and present the future work.

%% file: extraction.tex
\section{Extraction}%
\label{sec:extraction}

The aim of the extraction and annotation process was to build a corpus of at least 20,000 tweets that was as balanced as possible between the humor and not humor classes. Furthermore, as we intended to have a way of calculating the funniness score of a tweet, we needed to have several votes for the tweets that were considered humorous.

As we wanted to have both humorous and non-humorous tweet samples, we extracted tweets from selected accounts and from realtime samples. For the former, based on~\citet{castro2016joke}, we selected tweets from fifty humorous accounts from Spanish speaking countries, and took a random sample of size 12,000. For the latter, we fetched tweet samples written in Spanish throughout February 2018\footnote{The language detection feature is provided by the Twitter REST API.}, and from this collection we took another random sample of size 12,000. Note that we preferred to take realtime tweet samples as we did not want to bias by selecting certain negative examples, such as news or inspirational thoughts as in \citet{castro2016joke} and \citet{Mihalcea:2005:MCL:1220575.1220642}. From both sources we ignored retweets, responses, citations and tweets containing links, as we wanted the text to be self-contained. As expected, both sources contained a mix of humorous and non-humorous tweets. In the case of humorous accounts, this may be due to the fact that many tweets are used to increase the number of followers, expressing an opinion on a current event or supporting some popular cause. 

We first aimed to have five votes for each tweet, and to decide which tweet was humorous by simple majority. However, at a certain stage during the annotation process, we noticed that the users were voting too many tweets as non-humorous, and the result was highly unbalanced. Because of this, we made some adjustments in the corpus and the process: as the target was to have five votes for each tweet, we considered that the tweets that already had three non-humorous annotations at this stage should be considered as not humor, then we deprioritized them so the users could focus in annotating the rest of the tweets that were still ambiguous. We also injected 4,500 more tweets randomly extracted only from the humorous accounts. These new tweets were also prioritized since they had less annotations than the rest.


%% file: annotation.tex
\section{Annotation}%
\label{sec:annotation}

\begin{figure}
  \includegraphics[frame]{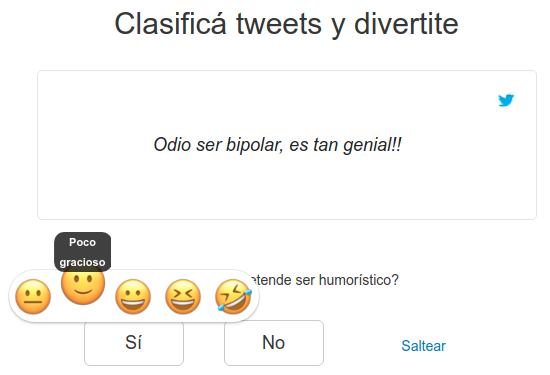}
  \caption{Example of a tweet presented to the annotators. It says: \emph{I hate being bipolar, it's so cool!}. The annotator is asked whether the tweet intends to be humorous. The available options are ``Yes'', ``No'' or ``Skip''. If the annotator selects ``Yes'', five emoji are shown so the annotator can specify how funny he considers the tweet. The emoji also include labels describing the funniness levels.}%
\label{fig:clasificahumor}
\end{figure}

A crowdsourced web annotation task was carried out to tag all tweets.\footnote{\url{https://clasificahumor.com}} The annotators were shown tweets as in \cref{fig:clasificahumor}. The tweets were randomly chosen but web session information was kept to avoid showing duplicates. We tried to keep the user interface as intuitive and self-explanatory as possible, trying not to induce any bias on users and letting them come up with their own definition of humor. The simple and friendly interface is meant to keep the users engaged and having fun while classifying tweets as humorous or not, and how funny they are, with as few instructions as possible.

If a person decides that a tweet is humorous, he has to rate it between one to five by using emoji. In this way, the annotator gives more information rather than just stating the tweet is humorous. We also allowed to skip a tweet or click a help button for more information. We consider that explicitly asking the annotator if the text intends to be humorous makes the distinction between the Not Humorous and Not Funny classes less ambiguous, which we believe was a problem of \citep{castro2016joke} user interface. Also, we consider our emoji rated funniness score to be clearer for annotators than their stars based rating.

The web page was shared on popular social networks along with some context about the task and the annotation period occurred between March 8\textsuperscript{th} and 27\textsuperscript{th}, 2018. The first tweets shown to every session were the same: three tweets for which we know a clear answer (one of them was humorous and the other two were not). These first tweets (``test tweets'') were meant as a way of introducing the user into how the interface works, and also as an initial way for evaluating the quality of the annotations. After the introductory tweets, the rest of the tweets were sampled randomly, starting with the ones with the least number of votes.

%% file: corpus.tex
\section{Corpus}%
\label{sec:corpus}

The dataset consists of two CSV files: tweets and annotations. The former contains the identifier and origin (which can be the realtime samples or the selected accounts) for each one of the \(27,282\) tweets\footnote{Tweet text is not included in the corpus due to Twitter Terms and Conditions. They can be obtained from the IDs.}, while the latter contains the tweet identifier, session identifier, date and annotation value for each one of the \(117,800\) annotations received during the annotation phase (including the times the skip button was pressed, \(2,959\) times). The dataset was released and it is available online.\footnote{\url{https://pln-fing-udelar.github.io/humor}}

When compiling the final version of the corpus, we considered the annotations of users that did not answer the first three tweets correctly as having lower quality. These sessions should not be used for training or testing machine learning algorithms. Fortunately, only a small number of annotations had to be discarded because of this reason. The final number of annotations is \(107,634\) (not including the times the skip button was pressed), including \(3,916\) annotations assigned to the test tweets themselves.

%% file: analysis.tex
\section{Analysis}%
\label{sec:analysis}


\subsection{Annotation Distribution}

\begin{figure}
  \definecolor{blue1}{rgb}{0.2,0.4,0.8}
  \begin{tikzpicture}
    \begin{axis}[xlabel=Number of annotations,ylabel=Tweets,area style,scale only axis,width=0.7\linewidth,height=65pt,minor y tick num=1,ymajorgrids,try min ticks=8,enlarge x limits=false,ymin=0,xtick style={draw=none},ytick pos=left]
      \addplot+[ybar,mark=no,blue1,blue1]
      table {data/histogram.dat};
    \end{axis}
  \end{tikzpicture}

  \caption{Distribution of tweets by number of annotations. Most tweets have between two and six annotations each.} 
\label{fig:histogram}
\end{figure}
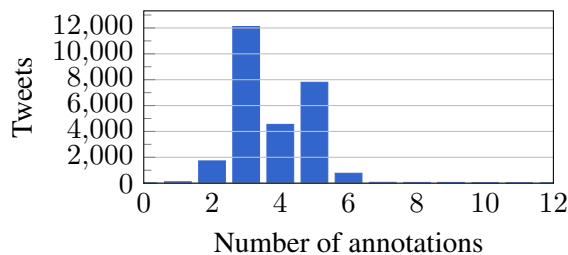

Each tweet received \(3.8\) annotations on average, with a standard deviation of \(1.16\), not considering the test tweets as they are outliers (they have a large number of annotations). The annotation distribution is shown in \cref{fig:histogram}. The histogram is highly concentrated: more than \(98\%\) of the tweets received between two and six annotations each. Even though the strategy was to show random tweets among the ones with less annotations, note that there are tweets with less than three annotations because some annotations were finally filtered out. At the same time, there are some tweets with more than six annotations because we merged annotations from a few dozen duplicate tweets. Also, note that there is a considerable amount of tweets with at least six annotations (\(1,001\)). This subset can be useful to study the different annotator opinions under the same instances.

\subsection{Class Distribution}%
\label{sec:class-distribution}

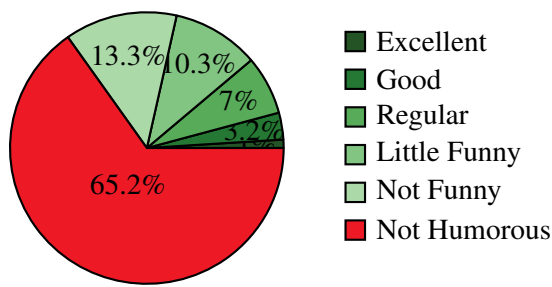
\begin{figure}
  \definecolor{green5}{rgb}{0.14,0.33,0.15}
  \definecolor{green4}{rgb}{0.14,0.47,0.2}
  \definecolor{green3}{rgb}{0.34,0.67,0.35}
  \definecolor{green2}{rgb}{0.51,0.77,0.51}
  \definecolor{green1}{rgb}{0.67,0.85,0.66}
  \definecolor{red1}{rgb}{0.93,0.1,0.15}
  \begin{tikzpicture}
    \pie[text=legend,color={green5,green4,green3,green2,green1,red1},radius=1.8]{1/Excellent, 3.2/Good, 7/Regular, 10.3/Little Funny, 13.3/Not Funny, 65.2/Not Humorous}
  \end{tikzpicture}
  
  \caption{Annotations according to their class.}%
\label{fig:annotations-by-tag}
\end{figure}

\cref{fig:annotations-by-tag} shows how the classes are distributed between the annotations. Roughly two thirds were assigned to the class Not Humorous, agreeing with the fact that there seem to be more non-humorous tweets from humorous accounts than the other way around. The graph also indicates that there is a bias towards bad jokes in humor, according to the annotators. We use simple majority of votes for categorizing between humorous or not humorous, and weighted average for computing the funniness score only for humorous tweets. The scale goes from one (Not Funny) to five (Excellent). Under this scheme, 27.01\% of the tweets are humorous, 70.6\% are not-humorous while 2.39\% is undecided (2.38\% tied and 0.01\% no annotations). At the same time, humorous tweets have little funniness overall: the funniness score average is \(1.35\) and standard deviation \(0.85\).

\subsection{Annotators Distribution}

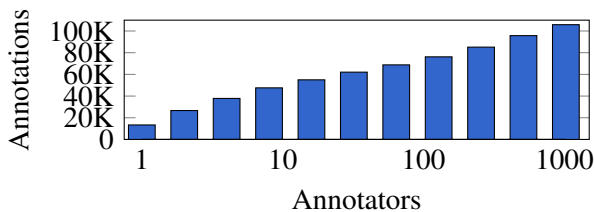
\begin{figure}
  \definecolor{blue1}{rgb}{0.2,0.4,0.8}
  \begin{tikzpicture}
    \begin{semilogxaxis}[
      xlabel=Annotators,
      ylabel=Annotations,
      width=\linewidth,
      height=90pt,
      xmin=0.75,
      xmax=1500,
      log origin=infty,
      enlargelimits=false,
      xtick style={draw=none},
      xtick={1,10,100,1000},
      xticklabels={1,10,100,1000},
      ymin=0,
      ymax=110000,
      ytick={0,20000,40000,60000,80000,100000},
      yticklabels={0,20K,40K,60K,80K,100K}
    ]
      \addplot[ybar, fill=blue1]
      table {data/accumulated-annotations.dat}
      \closedcycle;
    \end{semilogxaxis}
  \end{tikzpicture}

  \caption{Accumulated distribution of annotations by number of annotators. Notice that the top \(100\) annotators add up to more than \(70,000\) annotations.}%
\label{fig:annotators}
\end{figure}

There were \(1,271\) annotators who tagged the tweets roughly as follows: two annotators tagged \(13,000\) tweets, then one annotated \(8,000\), the next eight annotated between one and three thousand, the next \(105\) annotated between one hundred and one thousand and the rest annotated less than a hundred, having \(32,584\) annotations in total (see \cref{fig:annotators}). The average was \(83\) tags by annotator, with a standard deviation of \(597\).

\subsection{Annotators Agreement}

An important aspect to analyze is to what extent the annotators agree on which tweets are humorous. We used the \emph{alpha} measure from \citet{krippendorff2012content}, a generalized version of the \emph{kappa} measure \citep{cohen1960coefficient, fleiss1971measuring} that takes in account an arbitray number of raters. The agreement alpha value on humorous versus non-humorous is \(0.5710\). According to \citet{fleissstatistical}, it means that the agreement is somewhat between ``moderate'' to ``substantial'', suggesting there is acceptable agreement but the humans cannot completely agree. We believe that the carefully designed user interface impacted in the quality of the annotation, as unlike \citet{castro2016joke} this work's annotation web page presented less ambiguity between the class Not Humorous and Not Funny. We clearly outperformed their inter-annotator agreement (which was \(0.3654\)). Additionally, if we consider the whole corpus (including the removed annotations), this figure decreases to \(0.5512\). This shows that the test tweets were helpful to filter out low quality annotations.

Additionally, we can try to estimate to what extent the annotators agree on the funniness value of the tweets. In this case, disagreement between close values in the scale (e.g. Not Funny and Little Funny) should have less impact than disagreement between values that are further (e.g. Not Funny and Excellent). Following \citet{stevens1946theory}, in the previous case we were dealing with a \emph{nominal} measure while in this case it is an \emph{ordinal} measure. Alpha considers this into the formula by using a generic distance function between ratings, so we applied it and obtained a value of \(0.1625\) which is far from good; it is closer to a random annotation. There is a lack of agreement on the funniness. In this case, a machine will not be able to assign a unique value of funniness to a tweet, which makes sense with its subjectivity, albeit other techniques could be used~\citep{geng2016label}. In this case, if we consider the whole dataset, this number decreases to \(0.1442\).

If we only consider the eleven annotators who tagged more than a thousand times (who tagged \(50,939\) times in total), the humor and funniness agreement are respectively \(0.6345\) and \(0.2635\).




%% file: conclusion.tex
\section{Conclusion and Future Work}%
\label{sec:conclusion}

Our main contribution is a corpus of tweets in Spanish labeled by their humor value and funniness score with respect to a crowd-sourced annotation. The dataset contains \(27,282\) tweets coming from multiple sources, with \(107,634\) annotations. The corpus showed high quality because of the significant inter-annotator agreement value.

The dataset serves to build a Spanish humor classifier, but it also serves as a first step to tackle humor and funniness subjectivity. Even though more annotations per tweet would be appropriate, there is a subset of a thousand tweets with at least six annotations that could be used to study people's opinion on the same instances.

Future steps involve gathering more annotations per tweet for a considerable amount of tweets, so techniques such as the ones in \cite{geng2016label} could be used to study how people perceive the humorous pieces and what subjects and phrases they consider funnier. It would be interesting to consider social strata (e.g.\ origin, age and gender) when trying to find these patterns. Additionally, a similar dataset could be built for other languages which count with more data to cross over with (such as English) and build a humor classifier exploiting recent Deep Learning techniques based on it.